\documentclass[10pt,twocolumn,letterpaper]{article}

\usepackage{iccv}
\usepackage{times}
\usepackage{epsfig}
\usepackage{graphicx}
\usepackage{wrapfig}
\usepackage{amsmath}
\usepackage{amssymb}
\usepackage{float} 
\usepackage{multirow}

\usepackage{pdfpages}
\usepackage{multido}

\usepackage{listings}
\usepackage[pagebackref=true,breaklinks=true,letterpaper=true,colorlinks,bookmarks=false]{hyperref}

\usepackage{colortbl}

\iccvfinalcopy 

\def\iccvPaperID{4} 

\newcommand{\themetric}{CrossCoherence}
\newcommand{\notsosmall}{\fontsize{10.5pt}{12pt}\selectfont}

\ificcvfinal\fi

\begin{document}

\definecolor{gold}{rgb}{1.00, 0.84, 0.0}
\definecolor{silver}{rgb}{0.75, 0.75, 0.75}
\definecolor{bronze}{rgb}{0.8, 0.5, 0.2}

\title{Looking
 at words and points with attention:\\ a benchmark for text-to-shape coherence}

\author{ 
    Andrea Amaduzzi \hspace{1.5cm} Giuseppe Lisanti \hspace{1.5cm} Samuele Salti \hspace{1.5cm} Luigi Di Stefano \\ \\
    \notsosmall CVLAB, Department of Computer Science and Engineering (DISI)\\
    \notsosmall University of Bologna, Italy\\
    {\tt\small \{andrea.amaduzzi4, giuseppe.lisanti, samuele.salti, luigi.distefano\}@unibo.it}}

\maketitle
\ificcvfinal\thispagestyle{empty}\fi

\begin{abstract}
    While text-conditional 3D object generation and manipulation have seen rapid progress, the evaluation of coherence between generated 3D shapes and input textual descriptions lacks a clear benchmark. The reason is twofold: a) the low quality of the textual descriptions in the only publicly available dataset of text-shape pairs; b) the limited effectiveness of the metrics used to quantitatively assess such coherence. In this paper, we propose a comprehensive solution that addresses both weaknesses. Firstly, we employ large language models to automatically refine textual descriptions associated with shapes. Secondly, we propose a quantitative metric to assess text-to-shape coherence, through cross-attention mechanisms. To validate our approach, we conduct a user study and compare quantitatively our metric with existing ones. The refined dataset, the new metric and a set of text-shape pairs validated by the user study comprise a novel, fine-grained benchmark that we publicly release to foster research on text-to-shape coherence of text-conditioned 3D generative models. \\Benchmark available at  \normalsize\url{https://cvlab-unibo.github.io/CrossCoherence-Web/}.
    
\end{abstract}

\begin{figure}[t!]
    \centering
    \includegraphics[width=\linewidth]{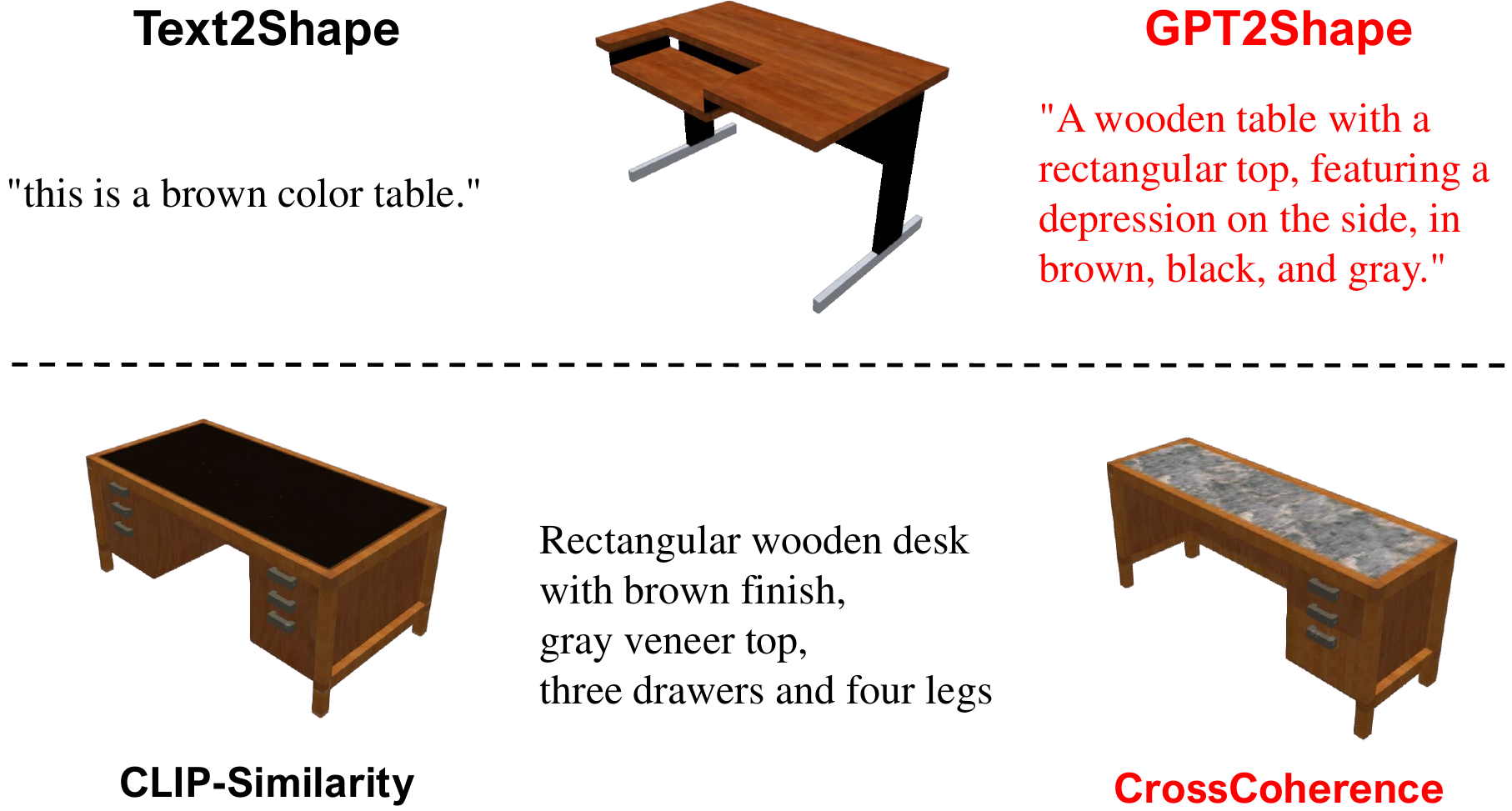}
    \caption{Benchmarking text-to-shape generative models calls for bett-er datasets and better metrics. \textbf{Top:} as existing  datasets~\cite{chen2019text2shape} contain many uninformative descriptions (left), we automatically create high-quality text prompts (right) by leveraging GPT-3~\cite{brown2020language}. \textbf{Bottom:} the existing metric CLIP-Similarity judges the left shape as more coherent to the given text than the right one, whilst our novel \themetric{} prefers the right shape over the left one.}
    \label{fig:teaser}
\end{figure}

\section{Introduction}
The rapid development of text-to-image generative models has enabled generation and manipulation of photo-realistic images from natural language prompts, in a matter of seconds~\cite{ramesh2021zero, ding2021cogview, nichol2021glide, ramesh2022hierarchical, gafni2022make, yu2022scaling, saharia2022photorealistic, feng2022ernie, balaji2022ediffi}. Inspired by such results, recent works~\cite{point_e,shap_e,liu2022towards, chen2019text2shape,poole2022dreamfusion,xu2022dream3d, sanghi2022clip,magic3d,fu2022shapecrafter} have started to explore the task of text-conditioned generation of 3D shapes. 

Two main paradigms are emerging for text-to-shape generation: one is a direct extension of the successful approach used to learn text-conditioned generative models for images and relies on supervised learning on high-quality paired text-shape
datasets~\cite{chen2019text2shape,liu2022towards,shap_e}. The extension of such paradigm to 3D shapes is however severely limited by the lack of general, high-quality 3D datasets: the only datasets for text-driven shape generation are Text2Shape~\cite{chen2019text2shape} and its extended version Text2Shape++~\cite{fu2022shapecrafter}, containing text descriptions for chairs and tables from ShapeNet~\cite{chang2015shapenet}. However, as shown in Figure~\ref{fig:teaser} (top, left), the textual descriptions provided by these datasets are often  generic and fail to capture all the key, fine-grained details of the objects in terms of both geometry and  appearance.
The second paradigm~\cite{sanghi2022clip, poole2022dreamfusion, magic3d} sidesteps the need for paired datasets by leveraging pre-trained CLIP~\cite{CLIP} or text-to-image models~\cite{ramesh2021zero, ding2021cogview, nichol2021glide, ramesh2022hierarchical, gafni2022make, yu2022scaling, saharia2022photorealistic, feng2022ernie, balaji2022ediffi} to learn to align 3D content creation to the input text. This paradigm has shown impressive results, but such methods are usually based on optimizing NeRF models~\cite{mildenhall2021nerf, mueller2022instant, dreamfields, poole2022dreamfusion, xu2022dream3d, magic3d} out of the generated images, which results in impractical run-time costs and latency (e.g., Magic3D~\cite{magic3d} takes 40 minutes on 8 NVIDIA A100 GPUs to generate one single shape).

Another important limitation for both paradigms is the lack of a common benchmark. It is impossible to overstate the importance of benchmarks in enabling and driving the rapid developments of deep learning: this is a cornerstone of the field since ImageNet~\cite{deng2009imagenet} gave rise to the AlexNet breakthrough~\cite{krizhevsky2017imagenet}. 
The definition of a clear benchmark for text-to-shape generation is hindered not only by the absence of high-quality paired text-3D datasets but also by metrics. Most currently adopted quantitative metrics can measure only the generative ability of a method, i.e. the realism of the generated shapes in terms of  distance between their distribution and the ground-truth one. These metrics, however, are not able to quantify how closely a 3D shape fits a text, arguably the most important aspect to properly evaluate text-to-shape methods. Exceptions are metrics based on CLIP embeddings~\cite{CLIP} which are computed from 2D renderings of the 3D shape and text embeddings. However, these approaches may turn out sensitive to rendering parameters and struggle to capture coherence with individual words or fine-grained geometric details due to the global nature of the CLIP text and visual embeddings. 



In this work, we propose the first benchmark for text-to-shape generation and manipulation that addresses both shortcomings highlighted above. Since collecting a novel dataset from scratch is a time-consuming and costly effort, we investigate on the effectiveness of pre-trained large language models (LLMs) to leverage existing datasets and improve their quality at a fraction of the cost. In particular, as shown in Figure~\ref{fig:teaser} (top, right), we create an improved version of Text2Shape, which we dub GPT2Shape, whose textual descriptions have been generated by the large language model GPT-3~\cite{brown2020language} starting from the text prompts of the original dataset. The higher quality of such text prompts has been validated through a user study. 
GPT2Shape provides multiple fine-grained textual descriptions for every 3D shape. 
The detail and accuracy of textual descriptions can enable training of generative models which can effectively learn the relationships between the words in the input text and the localized details of the corresponding 3D shape. 
Such capability would be very hard to achieve with noisy and imprecise text prompts. 
We also propose a novel metric, dubbed \themetric{}, which does not use renderings and exploits the cross-attention mechanism to quantify the coherence between a coloured point cloud and a textual description at the word and geometric detail level. We use the text-shape pairs validated through the user study to quantitatively compare with existing text-3D coherence metrics and show how \themetric{} outperforms previous proposals (Figure~\ref{fig:teaser}, bottom).  
Finally, as a by-product of our human evaluation study, we create the Human-validated Shape-Text (HST) dataset by collecting the textual descriptions which have been coherently associated with a shape by the participants. 
The usefulness of the HST dataset is twofold: it provides a set of prompts and associated shapes, that can be used as a test set of our benchmark since they are not present in the training set but come from the same distribution on which \themetric{} has been trained; it may also serve as a benchmark for the development of new text-to-shape coherence metrics.
 
To sum up, the main contributions of this work are:
\begin{itemize}
    \item \textbf{GPT2Shape}, an automatically improved dataset consisting of shape-text pairs with high-quality, fine-grained textual descriptions;
    \item \textbf{\themetric{}}, a state-of-the-art quantitative metric for text-to-shape coherence, which can be directly applied to RGB point clouds;
    \item \textbf{HST}, a human-validated test set where \themetric{} can be used to evaluate and compare text-driven generative models. 
\end{itemize}


\section{Related Work}
\label{related}
\textbf{Text-to-shape coherence metrics}:
Several recent studies have explored the problem of text-conditioned 3D generation. 
To assess fidelity to the input prompt,~\cite{liu2022towards}, \cite{sanghi2022clip}~and~\cite{chen2019text2shape} compute several distance measures between a generated shape and the one associated with the input prompt in the ground-truth set, such as Intersection-over-Union (IoU), Earth's Mover Distance (EMD), Chamfer Distance (CD) and Mean Squared Error (MSE). The main limitation of such a strategy is that, in generative tasks, a ground-truth shape is just one possible correct outcome of the generation process and relatively higher or lower distances from it do not measure the quality of the output nor, in the case of text-conditioned generation, its coherence with the given text. 
The most convincing metrics for text-to-shape coherence are based on CLIP embeddings~\cite{CLIP}. One such metric is CLIP R-precision~\cite{park21clip_r_precision}, used in several studies~\cite{point_e,liu2022towards,poole2022dreamfusion,xu2022dream3d, shap_e}. It is defined as the accuracy with which CLIP retrieves the correct caption among a set of distractors given a rendering of the generated shape. Yet, such quantitative assessment is affected by several limitations: first of all, it is based on renderings, therefore it is influenced by a number of parameters, like virtual camera placement, virtual illumination, and number of rendered views; secondly, the selection of the text distractors is arbitrary; finally, as it relies purely on rendered views, whose quality is not homogeneous among 3D data representations, a comparison between 3D shapes from different representations (e.g. point clouds and meshes) can be unfair.
CLIP-similarity~\cite{fu2022shapecrafter}, instead, computes the cosine-similarity between CLIP image features extracted from several shape renderings and CLIP text features. Due to the reliance on renderings, this metric shares the same limitations of CLIP R-precision.

ShapeCrafter~\cite{fu2022shapecrafter} has been the first and only work to use ShapeGlot~\cite{achlioptas2019shapeglot} as a text-shape fidelity metric. In its original formulation, given a set of shapes, this metric is able to discriminate which one is the most coherent with respect to an input text description, by extracting 3D features, 2D features from  renderings, and text features. 
Despite its usefulness, there are some clear disadvantages associated with this method. Firstly, similar to CLIP-based metrics, its effectiveness relies on the parameter settings used to render the views of the shape. Secondly, the text representation employs an LSTM architecture, which has been acknowledged to have difficulties in representing long  text descriptions~\cite{bengio94}.

To address all the above limitations, we propose \themetric{}, a metric for text-to-shape coherence that does not depend on rendering parameters, operates directly on colored point clouds to take both geometry and appearance into account, leverages LLMs to extract better text features and exploits cross-attention to assess the coherence of words with shape parts.

\textbf{Datasets} Several text-driven 3D generation models require supervision in the form of text-shape pairs. 
Unlike in the 2D case, where web scraping can be used to obtain large amounts of image-text pairs~\cite{coco_captions_2015, thomee2016yfcc100m, jft300m2017, schuhmann2022laion}, obtaining such labeled data is extremely difficult, due to 3D shapes not being common on the Internet and the complexity and cost of creating text descriptions for large 3D datasets like ShapeNet~\cite{chang2015shapenet}. 
Text2Shape~\cite{chen2019text2shape} has been the first dataset with shape-text pairs to be introduced. 
It contains text prompts for two Shapenet classes: chairs and tables. 
Overall, Text2Shape provides 75K shape-text pairs, with  multiple text descriptions for each 3D object. 
However, the quality of these descriptions is highly variable: the prompts are either too generic (\emph{``A chair with four legs''}) or contain irrelevant details (\emph{``Table with high-quality wood, you can eat and enjoy with your family''}) or both. Sometimes the text does not describe correctly the geometry and/or appearance of the objects (\emph{``A table with three legs''} but the table has actually four legs). 
In addition, many sentences contain grammatical and syntactical errors. 
More examples are provided in the supplemental. 
These nuances make the process of learning the complex relationship between 3D data and text even more difficult and limit the validity of comparisons carried out on this dataset.
Text2Shape++~\cite{fu2022shapecrafter} is a dataset built upon Text2Shape, including 369K shape-text pairs. This dataset has been built specifically for the task of recursive shape generation; indeed, the original text descriptions from Text2Shape have been split into multiple smaller sentences with incremental amount of information (e.g. \emph{``a chair''}, \emph{``a chair with four legs''}, \emph{``a chair with four legs and no armrests''}). Since Text2Shape++ is built starting from the same descriptions of Text2Shape, its text prompts suffer from the same limitations.  
Objaverse~\cite{deitke2023objaverse} is a very recent large-scale dataset that provides more than 800K 3D models together with descriptions. However, the majority of these texts specify only the class of the object and some attributes, without conveying information about its actual 3D geometry and appearance. 

In response to these challenges, we propose an automatically improved version of Text2Shape, referred to as GPT2Shape, which contains new text descriptions for every object of Text2Shape. These novel sentences can describe more accurately the geometry, appearance and texture of all the objects in the dataset, as proved by our human evaluation study and exemplified in Figure~\ref{fig_gpt3}.




\section{GPT2Shape}
\begin{figure}[!ht]
\centering
   \includegraphics[width=0.95\columnwidth]{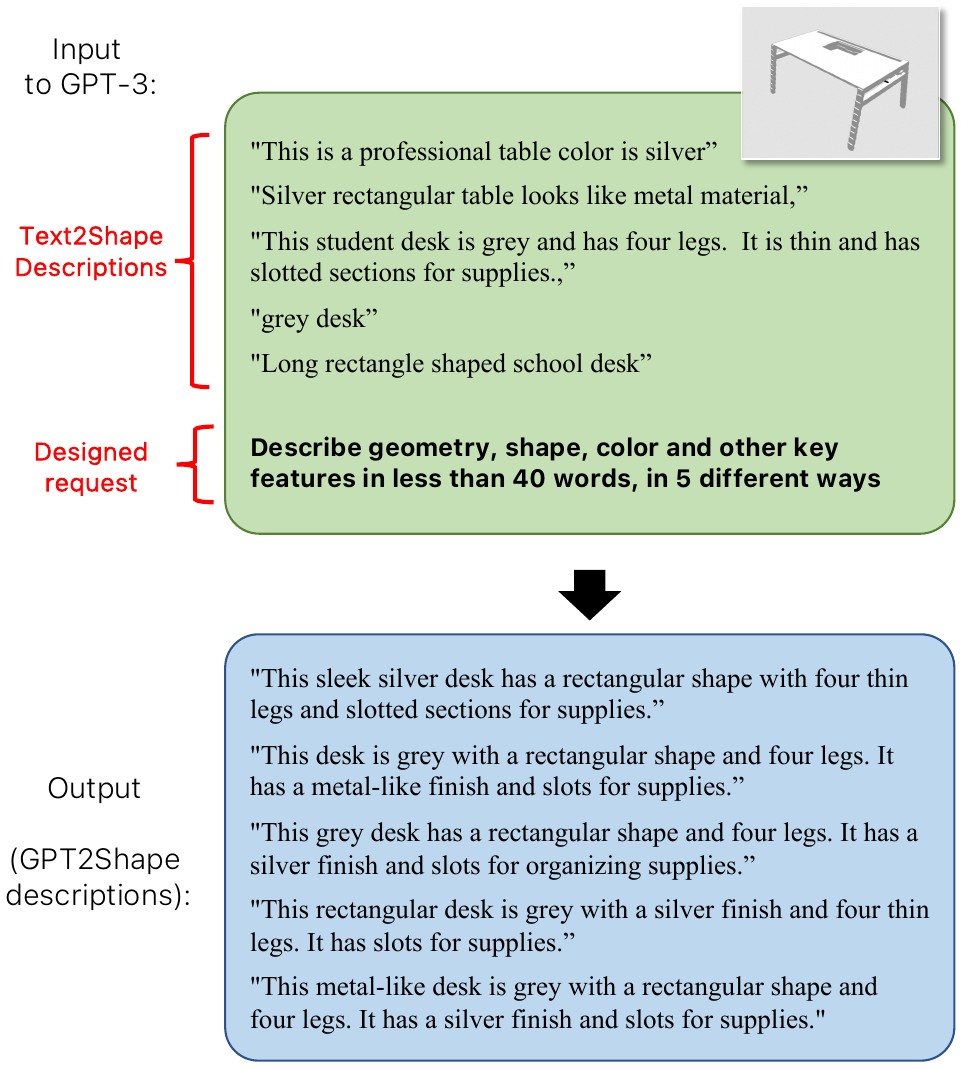}
   \caption{An example of the text prompts rephrasing process. The green box contains Text2Shape descriptions for a shape together with the designed request; their combination represents the input prompt to GPT-3. The blue box is the output of GPT-3, providing the textual descriptions of GPT2Shape.}
   \label{fig_gpt3}
\end{figure}

Figure~\ref{fig_gpt3} presents an exemplar shape and its prompts from Text2Shape. As it can be seen from this figure, the quality of the original descriptions is highly variable and some of them are quite poor: for instance, the sentence \emph{``grey desk''} 
is too generic and not able to drive any generative model towards the intended 3D shape. 
Improving such dataset through a new human-based annotation on the Amazon Mechanical Turk crowd-sourcing platform~\cite{crowston2012amazon}, utilized by Text2Shape, would be extremely time-consuming and expensive. \\

At the same time, enough details about a shape are usually present in Text2Shape if information from multiple prompts are merged, as shown again in Figure~\ref{fig_gpt3}. On the basis of this observation, we decided to explore if an automatic improvement of the existing dataset was feasible, by exploiting LLMs. In particular, the generation of the new text prompts has been achieved through the OpenAI GPT-3 model~\cite{brown2020language} \textit{davinci}. 
GPT-3 has been used in its text completion scenario in this manner: for each shape of the dataset, all its corresponding text prompts from Text2Shape have been given as input to the language model, along with the request to generate better text descriptions.
The request itself was the result of a careful prompt engineering process, in order to obtain the best possible result. Figure~\ref{fig_gpt3} illustrates an example of this process, showing the initial text descriptions, the request to GPT-3, and the resulting sentences. 
As it can be noticed, the output textual descriptions contain all the relevant geometric and appearance information regarding the object. 
This text rephrasing process was carried out on the whole train-val-test splits of Text2Shape, so as to keep the same amount of data samples as the original dataset: $15032$ shapes, with a total of $75358$ shape-text pairs. 
In the supplementary material, multiple comparisons between text prompts from Text2Shape and GPT2Shape are provided.

\section{GPT2Shape vs Text2Shape: a user study}
\label{sec:humaneval}
\begin{figure}[t!]
\centering
   \includegraphics[width=0.8\linewidth]{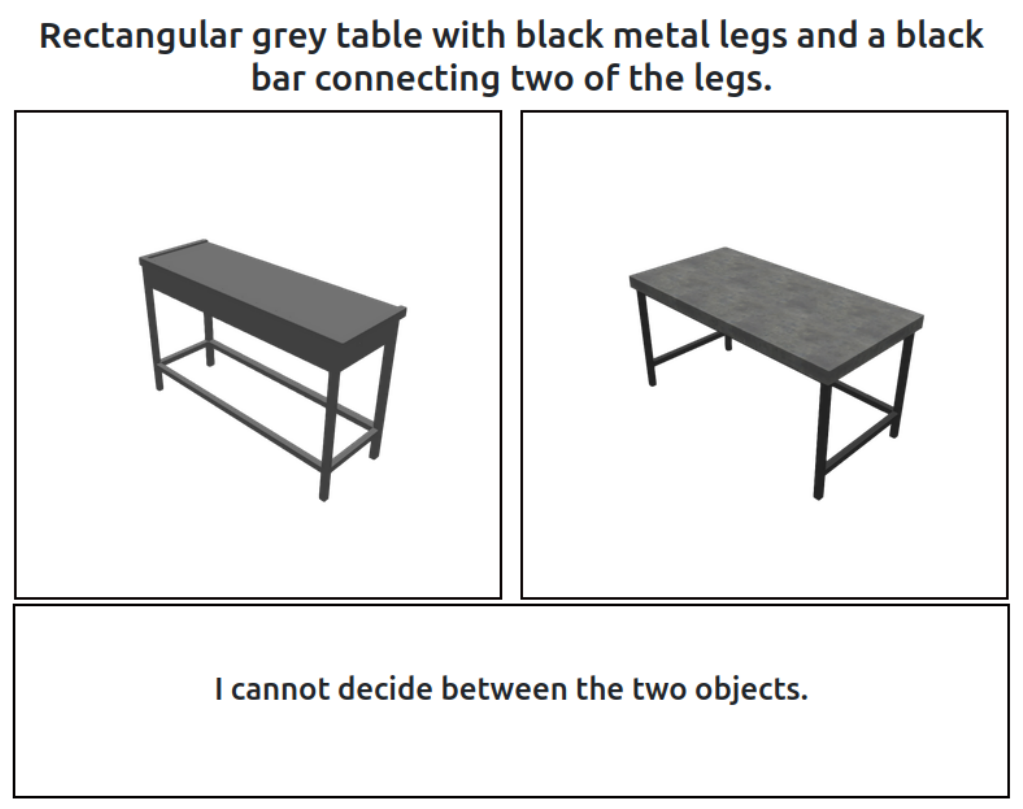}
    \caption{Layout of the user study}
   \label{fig_humaneval}
\end{figure}

Once the rephrased sentences had been obtained via GPT-3, a user study aimed at comparing the informativeness of the old (Text2Shape) and new (GPT2Shape) text prompts was carried out.  
The layout of this user study is shown in Figure~\ref{fig_humaneval}: each user was shown a pair of views from two objects together with a text prompt. The prompt comes from the test set of either Text2Shape or GPT2Shape and describes one of the two objects (i.e. the reference object, unknown to the user) while the other one acts as a distractor. The left/right position of the reference object was randomized. The task  consisted in clicking on the object which, according to the user, would be described better by the given text. 
In case of uncertainty, due to the prompt describing equivalently well both objects or not sufficiently well any of the two, the user could click on a third option, located under the images. 
This enabled us to compare the two datasets based on users' choices:  the higher the number of clicks on the -unknown- reference object, the higher the informativeness of the associated text prompt.


To avoid presenting the user with a task far too simple, as it might have been the case had distractors been chosen randomly, the \emph{reference-distractor-text} triplets were built as follows. For every reference object, we have defined two \textit{hard} distractors and one \textit{easy} distractor. All distractors for a shape have been selected within its class (e.g., chairs or tables) and defined on the basis of the Euclidean distance in the latent space of a PointNet++~\cite{qi2017pointnet++} autoencoder trained on the task of RGB point cloud reconstruction on chairs and tables from
ShapeNet.
In this way, we obtained three pairs for each reference object. When running the user study, a pair of shapes were randomly sampled from this set, whereas the text describing the reference shape was taken from Text2Shape or GPT2Shape with $50\%$ probability. 
In the supplemental, we provide more details and examples on easy-hard shape distractors and triplets from the user study.

Overall, $175$ users took part in the study, contributing with $3642$ data samples (i.e., clicks). In order to evaluate Text2Shape and GPT2Shape, we kept track of the answers provided by the users when presented with text prompts from both datasets. Table~\ref{table_humaneval} summarizes the results by reporting the percentages of times users selected the reference object, the distractor, or felt they could not decide when presented with a text from Text2Shape or GPT2Shape. When the text came from GPT2Shape, about $75\%$ of the answers were correct, i.e. the human selected the reference shape as the most coherent with the provided text description. 
On the other hand, Text2Shape allowed the users to answer correctly only $65\%$ of the time. 
Moreover, the percentage of undecidable sentences is higher for Text2Shape (about $25\%$) compared to GPT2Shape (about $16\%$).

Such experimental findings validate our claim that large language models can be used to automatically filter and combine information from multiple incomplete textual descriptions. 
They also validate that textual descriptions in GPT2Shape are more informative than the original prompts from Text2Shape.

\begin{table}[h]
\centering
\scalebox{0.85}{
\begin{tabular}{|c|c|c|c|}
\hline
Dataset & reference ($\uparrow$) & cannot decide ($\downarrow$) & distractor ($\downarrow$)\\
\hline
\textbf{Text2Shape} & 65.65\% & 24.95\% & 9.40\% \\
\textbf{GPT2Shape} & \textbf{75.76}\% & \textbf{16.21}\% & \textbf{8.02}\% \\
\hline
\end{tabular}
}
\caption{Summary of user study results}
\label{table_humaneval}
\end{table}

\section{Human-validated Shape-Text dataset}\label{sec:test_set}
The results of the user study were used also to aggregate a refined test set of shape-text pairs with descriptive texts from the test sets of Text2Shape and GPT2Shape, which we dub Human-validated Shape-Text (HST) dataset. When building this subset, we only selected the sentences for which all users made the same choice between the two objects, i.e., where there was unambiguous consensus among participants that the description was describing well only one of the two shapes. 
HST, which contains $2153$ text-shape pairs, provides a human-validated benchmark to assess the quality of text-to-shape generative models. It can be used as a set of descriptive text prompts to evaluate such models with text-to-shape coherence metrics.
To provide a comprehensive dataset, for every pair in the HST dataset, we also included the other object that was shown to the human evaluators, along with the indication of the correct association for the text. 
This extended version of HST is useful for quantitatively validating metrics for text-to-shape coherence.

\section{\themetric{} \label{sec:themetric}}
In this section, we describe the details of our proposed metric for text-to-shape coherence, \themetric{}. 
Recent works on text-conditioned generative models, both in 2D~\cite{ramesh2021zero, ding2021cogview, nichol2021glide, ramesh2022hierarchical, gafni2022make, yu2022scaling, saharia2022photorealistic, feng2022ernie, balaji2022ediffi} and 3D~\cite{magic3d, xu2022dream3d, point_e, liu2022towards, shap_e, poole2022dreamfusion, fu2022shapecrafter} have shown that frozen large language models, like BERT~\cite{devlin2018bert}, GPT~\cite{brown2020language} and T5~\cite{t5}, trained only on text data, can be very effective text encoders for generative tasks.
The solution proposed in~\cite{imagen} has also shown that large language models are more effective than text encoders trained on paired image-text data, such as CLIP~\cite{CLIP}. 
Another key reason for the outstanding results achieved by such works is the text-conditioning scheme, which is based on the cross-attention mechanism. 
On the basis of such observations, we have decided to explore the use of large language models and cross-attention layers for the definition of our metric.

\subsection{Architecture}
\begin{figure}[t]
\centering
   \includegraphics[width=\linewidth]{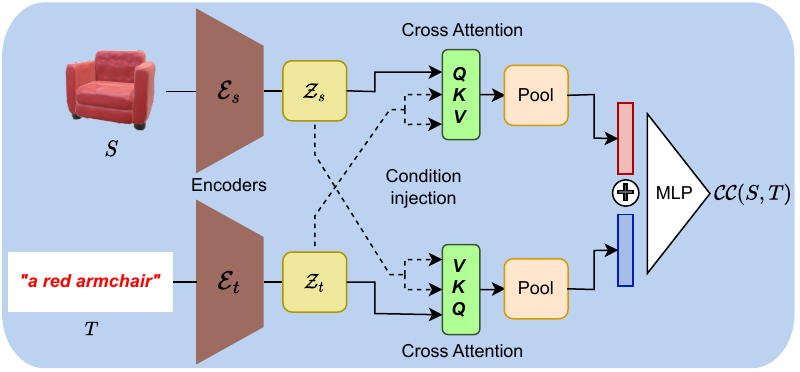}
   \caption{Architecture of \themetric{} \label{fig:metric}}
\end{figure}

The proposed architecture of \themetric{} is shown in Figure~\ref{fig:metric}. 
Our model takes as input a text description and a group of colored point clouds, with coordinates \textit{(x, y, z)} as well as \textit{(R, G, B)} values for every point. 
It provides as output a set of soft-max normalized scores, which are used to predict which point cloud is best described by the input text. 
In practice, a Point Cloud Encoder extracts meaningful local embeddings from every input point cloud. 
At the same time, a Text Encoder extracts text features from the input sentence. 
Then, for every point cloud, two \textit{bilateral}  cross-attention layers reason on the coherence between the shape and text features computed by the encoders. These layers are referred to  as \textit{bilateral} since, in a first layer, the queries of the attention maps are computed on the shape features, whereas the keys and values are obtained from the text embeddings, while in a second layer this computation is reversed. 
This \textit{bilateral} layer equips the metric with the ability to reason on the mutual relation between words and 3D details, which is key to evaluating text-to-shape coherence, as shown by the experiments reported in Section~\ref{sec:expres}. Finally, the resulting embeddings from both cross-attentions are concatenated, after undergoing an average pooling layer. These features are fed into an MLP in order to provide a score of coherence for each point cloud and text. At training time, the scores are softmax-normalized and a standard cross-entropy loss is used to guide the network to produce higher scores for the correct pair.
As Point cloud Encoder we used the encoder part of a PointNet++ network~\cite{qi2017pointnet++} trained for Shape Reconstruction on chairs and tables from ShapeNet. This network is the same model used to build the easy and hard distractors for the user study described in Section~\ref{sec:humaneval}.
When building the distractor pairs, we used the features coming from the third set abstraction layer, which computes a global embedding, whereas when training \themetric{} we used the embeddings computed by the second set abstraction layer, which computes local features on the downsampled input point cloud. 
As for the Language Encoder, we used the frozen encoder from T5~\cite{t5}. 


\subsection{Training}
\label{train_attnglot}
During training, the network was fed with  \textit{(shapes, text}) pairs and   target vectors indicating  the ground-truth shape corresponding to the textual description. 
Based on the outcome of the user study, we trained \themetric{} on the training set of GPT2Shape. 
Before training, following the same strategy applied for the human evaluation survey, described in Section~\ref{sec:humaneval}, for every shape in the training set of GPT2Shape, we have extracted one easy and two hard distractors. 
This \textit{easy-hard} training strategy was found to be more effective than constructing the pairs randomly due to the network learning to deal with cases where the input point clouds are very similar to each other, which in turn is conducive to better  generalization to unseen examples. 


\section{Experimental Results \label{sec:expres}}
We performed several experiments to assess the effectiveness of \themetric{}. 

\begin{table}[t!]
\centering
\scalebox{0.85}{
\begin{tabular}{|l|c|c|}
\hline
\multicolumn{1}{|p{2cm}|}{\textbf{Method}} & 
\multicolumn{1}{|p{2.5cm}|}{\textbf{Acc.} on chairs ($\uparrow$)} & 
\multicolumn{1}{|p{2.9cm}|}{\textbf{Acc.} on full HST ($\uparrow$)} \\
\hline
ShapeGlot & 70.21\% & - \\
\hline
CLIP-Similarity & 77.79\% & 77.06\% \\
\hline
\themetric{} & \textbf{81.04}\% & \textbf{80.45}\% \\
\hline
\end{tabular}
}
\caption{Comparison between ShapeGlot, \themetric{} and CLIP-Similarity. The second column reports accuracy for methods trained and tested only on chairs, while the third column is for methods trained and tested on full HST.}
\label{tab:HTS}
\end{table}

\begin{table}[t!]
\centering
\scalebox{0.85}{
\begin{tabular}{|l|c|c|c|}
\hline
\textbf{Test set} & ShapeGlot & \themetric{} & CLIP R-precision \\
\hline
HST chairs & 5.36\% & \textbf{17.26}\% & 9.87\%\\
\hline
full HST & - & \textbf{16.85}\% & 9.38\% \\
\hline
\end{tabular}
}
\caption{Comparison between ShapeGlot, \themetric{} and CLIP R-precision based on the R-precision protocol. The second row reports the results for methods trained and tested only on chairs, while the third row is for methods trained and tested on full HST.}
\label{tab:HTS_Rprec}
\end{table}

\subsection{Comparison with existing metrics \label{subsec:expres_HST}}

In this section, we evaluate the accuracy in assessing text-to-shape coherence for our proposed \themetric{} metric, as well as for Shapeglot~\cite{achlioptas2019shapeglot}, CLIP-Similarity~\cite{fu2022shapecrafter, CLIP} and CLIP R-precision~\cite{park21clip_r_precision,  CLIP}.
The accuracy computation relies on the triplets of HST (Section~\ref{sec:test_set}) and consists in two different evaluation protocols. 
When comparing with CLIP-Similarity, we report how often each metric yields a higher score to the ground truth shape than to the other one in the triplet for the given text description. 
In contrast, when dealing with CLIP R-precision, we adopt the protocol introduced in~\cite{dreamfields} and used by~\cite{point_e,liu2022towards,poole2022dreamfusion,xu2022dream3d}: given a text prompt and the corresponding ground-truth 3D shape from HST, we construct a set of 153 text descriptions. 
This set comprises the ground-truth prompt and 152 other texts randomly sampled from a designated test set, in our case HST. 
Every metric will predict which text prompt within the set  exhibits the highest coherence to the given 3D shape. 
By comparing the predicted prompt to the ground-truth prompt, we can assess the performance of the evaluation metric.
To evaluate both CLIP-Similarity and CLIP R-precision, we conducted experiments using various CLIP base models and different numbers of renderings.  
When dealing with more than one rendered view, we used the maximum similarity across the views, as in~\cite{fu2022shapecrafter}. 
The configuration that achieved the best performance, which we consider here,  relies on ViT-L/14 Encoder and 20 renderings.
Yet, we deem it worth highlighting that the accuracy of CLIP-based metrics varies significantly depending on the chosen configuration (CLIP base model, 3D data representation, rendering parameters), as shown in additional experiments reported in the supplementary material.
As for ShapeGlot~\cite{achlioptas2019shapeglot}, we trained the network on the same dataset as \themetric{}, i.e. easy and hard samples from GPT2Shape, as described in Section~\ref{sec:themetric}. However, while training \themetric{} relies only on point clouds and text prompts, ShapeGlot consumes also rendered views of objects' meshes and requires a pre-trained VGG encoder to extract features from these renderings. Unfortunately, the authors did not release the weights of the VGG model and made available only pre-computed features for the \textit{chair} class of Text2Shape. Hence, we can assess the accuracy of ShapeGlot only on the chair subset of HST.

\begin{figure}[t!]
    \centering
    \includegraphics[width=0.89\linewidth]{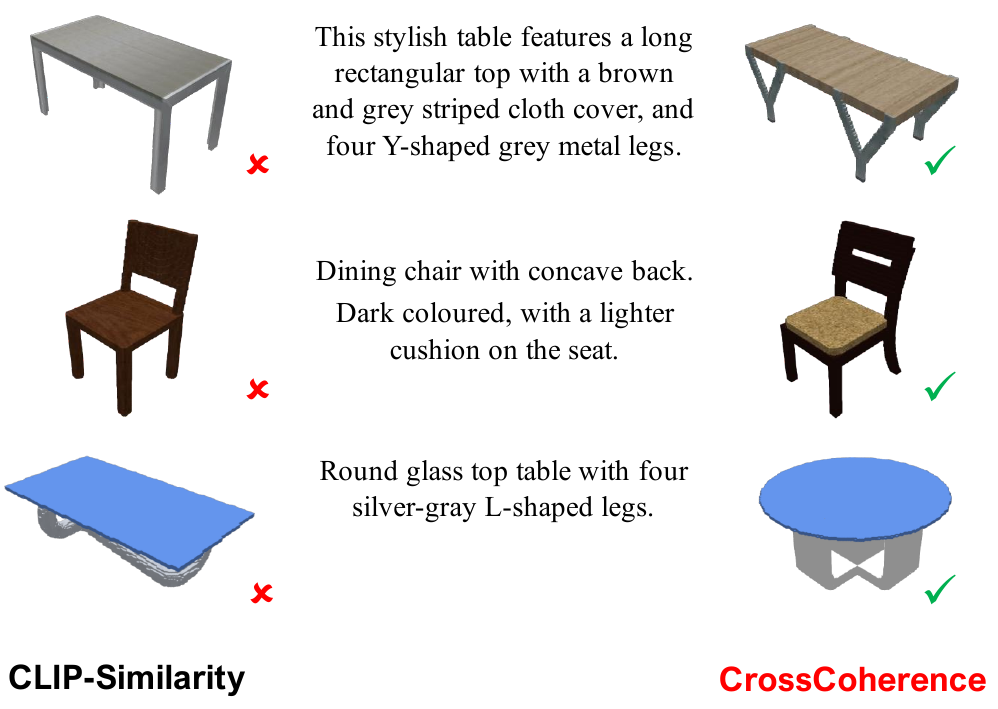}
    \caption{Qualitative results of CLIP-Similarity and \themetric{} on HST dataset. The green check indicates the shape associated with the prompt, while the red cross identifies the distractor. For all these triplets, CLIP-Similarity prefers the left shape while \themetric{} the right one.}
    \label{fig:quali_HTS}
\end{figure}

\begin{figure}[t!]
    \centering
    \includegraphics[width=0.89\linewidth]{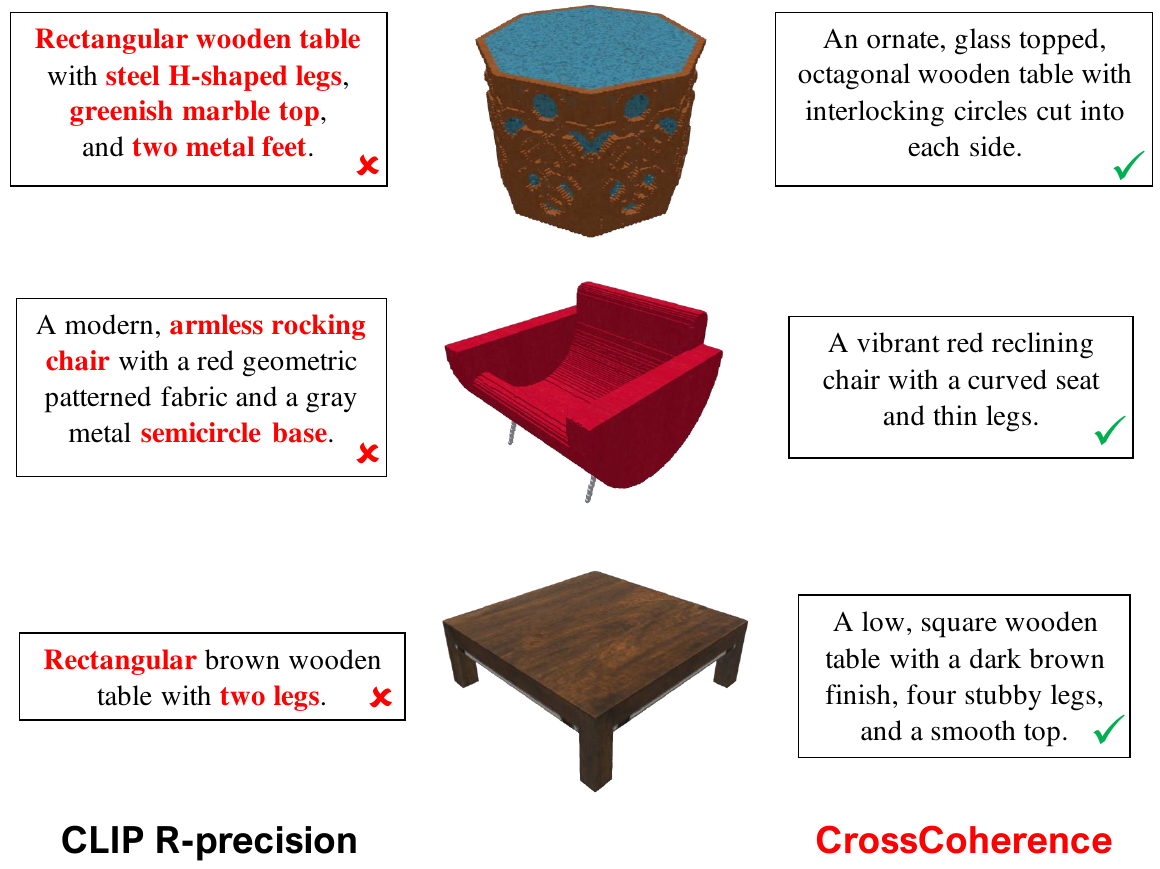}
    \caption{Qualitative results of CLIP R-precision and \themetric{} on HST dataset. The green check indicates the ground-truth text. For all these triplets, CLIP R-precision retrieves the left text while \themetric{} the right one.}
    \label{fig:quali_HTS_Rprec}
\end{figure}

The outcome of our evaluation is shown in Tables~\ref{tab:HTS} and~\ref{tab:HTS_Rprec}. 
These results indicate that \themetric{} is the most effective metric. 
We can observe how ShapeGlot performance is the least satisfactory. Moreover, the gap in accuracy between this metric and \themetric{} suggests that modern text-analysis tools like LLMs and cross-attention layers are key to creating a more effective metric. Indeed, the main differences between \themetric{} and ShapeGlot include the use of LLMs instead of LSTMs  as  well as the incorporation of the cross-attention mechanism. 
Finally, in both settings, CLIP-based metrics turn out to be the strongest competitors; however,  \themetric{} outperforms CLIP-Similarity by more than $3\%$ and is almost $100\%$ more accurate than CLIP in the challenging R-precision setting, which is nowadays the most widely used protocol for evaluating text-to-shape coherence.


Qualitative examples of the choices made by CLIP-Similarity and \themetric{} are provided in Figure~\ref{fig:quali_HTS}. 
The errors made by CLIP-Similarity seem to be caused by the use of global embeddings to compare shapes and text, which prevents the metric from evaluating fine details when judging the shape-text agreement. 
For instance, CLIP-Similarity does not take into account the shape of the table legs in the first row, the presence of the chair cushion in the second row and the geometry of the glass table in the last example.
As shown in Figure~\ref{fig:quali_HTS_Rprec}, CLIP R-precision suffers from the same weakness as CLIP-Similarity in understanding fine-grained details of visual input and text. For example, CLIP R-precision provides high scores for text prompts that contain multiple wrong references to the geometry and appearance of the input shape, i.e.\emph{``Rectangular wooden table''} in the first row, \emph{``Armless rocking chair''} in the second example, \emph{``two legs''} in the last row.
This weakness has been highlighted in~\cite{beyer23} and~\cite{aro}, which discuss how CLIP models, probably due to their contrastive pretraining strategy, have difficulty in processing fine-grained descriptions of visual content. Indeed, the performance of such models on the Attribution, Relation, and Order (ARO) benchmark~\cite{aro} show that CLIP exhibits poor relational understanding and tends to behave like bag-of-words models.  
The use of cross-attention between local embeddings in \themetric{}, on the other hand, enables to capture and link together important shape and text details, leading to the selection of the reference shape or text, as shown in Figures~\ref{fig:quali_HTS} and~\ref{fig:quali_HTS_Rprec}, respectively.
Ablation experiments concerning the design choices behind our architecture are reported in the supplementary material, together with additional qualitative comparisons between the metrics.

\begin{table}[t!]
\centering
\scalebox{0.74}{
\begin{tabular}{|l|c|c|c|}
\hline
\multicolumn{1}{|p{2.0cm}|}{\textbf{Method}} & 
\multicolumn{1}{|p{2.3cm}|}{\centering \textbf{{\themetric{}}}} & 
\multicolumn{1}{|p{2.5cm}|}{\centering \textbf{CLIP-Similarity}} & 
\multicolumn{1}{|p{2.5cm}|}{\centering \textbf{CLIP-Similarity}}\\
\hline
Point-E \cite{point_e} & \cellcolor{bronze}27.33\% & \cellcolor{silver}24.36\% & \cellcolor{silver}27.04\%\\
\hline
Shap-E \cite{shap_e} & \cellcolor{gold}\textbf{38.02}\% & \cellcolor{gold}\textbf{65.47}\% (STF) & \cellcolor{gold}\textbf{57.18}\% (NeRF)\\
\hline
Liu et al. \cite{liu2022towards} & \cellcolor{silver}34.65\% & \cellcolor{bronze}10.18\% & \cellcolor{bronze}15.78\%\\
\hline
\end{tabular}
}
\caption{Quantitative comparison between generative models, using \themetric{} and CLIP-Similarity. Each entry in a column reports how often a metric considers a prompt more coherent with the shape generated by one method versus the others. \label{tab:cc_clipsim}}
\end{table}

\begin{table}[t!]
\centering
\scalebox{0.74}{
\begin{tabular}{|l|c|c|c|}
\hline
\multicolumn{1}{|p{2.3cm}|}{\textbf{Method}} & 
\multicolumn{1}{|p{2.7cm}|}{\textbf{{CrossCoherence}}} & 
\multicolumn{2}{|c|}{\textbf{CLIP R-precision}} \\
\hline
Point-E \cite{point_e} & \cellcolor{bronze}4.92\% & 
\multicolumn{2}{|c|}{\cellcolor{silver}5.95\%} \\
\hline
Shap-E \cite{shap_e} & \cellcolor{gold}\textbf{7.09}\% & \cellcolor{gold}\textbf{20.41}\% (STF) & \cellcolor{gold}\textbf{15.78}\% (NeRF)\\
\hline
Liu et al. \cite{liu2022towards} & \cellcolor{silver}5.09\% & 
\multicolumn{2}{|c|}{\cellcolor{bronze}5.83\%} \\
\hline
\end{tabular}
}
\caption{Quantitative comparison between generative models, using \themetric{}  and CLIP R-precision according to  the R-precision protocol. 
    \label{tab:cc_rprecision}}
\end{table}

\subsection{Text-to-shape coherence of generative methods}
\label{sec:gen_results}
A text-to-shape coherence metric can be used to evaluate text-conditioned 3D shape generative models. Here we address this setting and compare \themetric{}, CLIP-Similarity~\cite{fu2022shapecrafter} and CLIP R-precision~\cite{park21clip_r_precision} in their ability to judge upon the coherence to the input text of the shapes  generated by \textit{Point-E}~\cite{point_e}, \textit{Shap-E}~\cite{shap_e} and the model proposed by Liu et al.~\cite{liu2022towards}.
These models employ distinct data representations, with Point-E being the current state-of-the-art method in text-driven point cloud generation, Shap-E using both NeRF (neural radiance field) and STF (signed textured field), Liu et al.~\cite{liu2022towards} employing occupancy fields.

We used the pretrained weights provided by the authors for Point-E and Shap-E, while we trained from scratch Liu et al.~\cite{liu2022towards} on GPT2Shape. 
We then used CLIP R-Precision, CLIP-Similarity and \themetric{} to assess which method generates the 3D data most faithful to the input prompt. 
We adopted the same experimental protocols as in Section~\ref{subsec:expres_HST}, the only difference being that here the 3D shapes compared to the text prompts are not the ground-truths from HST but have been generated by either Shap-E or Point-E or Liu et al.

\begin{figure}[t!]
    \centering
    \includegraphics[width=0.99\linewidth]{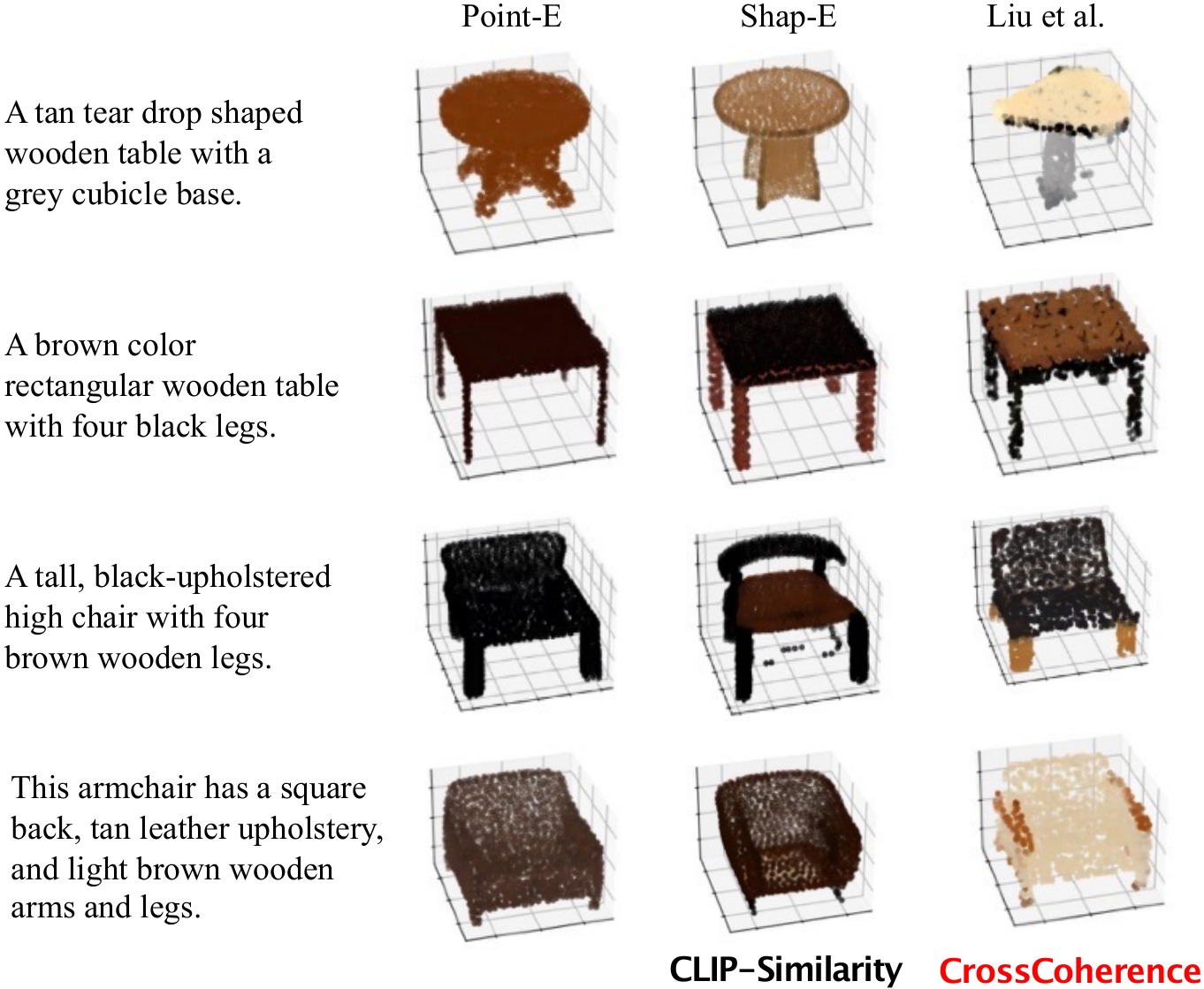}
    \caption{Examples of shapes generated by Point-E  (left), Shap-E (center) and Liu et al. (right) for the given text prompt. As highlighted, for all these triplets, CLIP-Similarity prefers the shape in the middle while \themetric{} the one on the right.}
    \label{fig:quali_cc_clipsim}
\end{figure}

Results are reported in Tables~\ref{tab:cc_clipsim} and~\ref{tab:cc_rprecision}.
Since Shap-E represents the generated shape as an STF and a NeRF, for CLIP-Similarity and CLIP R-precision we report the results  computed on renderings from both representations. The differences in performance for Shap-E  suggest again that CLIP-based metrics are significantly affected by the rendering process. 
In these experiments, we notice that all metrics agree on Shap-E being the method that generates the 3D shapes closest to the input text. 
However, the ranking of the other two methods in relation to Shap-E is very different.  In fact, \themetric{} ranks the method of Liu et al.~\cite{liu2022towards} second, with a modest gap of 4\% in the first protocol and 2\% in the second. 
In contrast, CLIP-similarity and CLIP R-precision rank  Point-E in second place with a very large gap in performance with respect to Shap-E (around 30\% in the first protocol, and from 10\% to 15\% in the second one).
By inspecting  some triplets on which \themetric{} and CLIP-Similarity disagree, shown in Figure~\ref{fig:quali_cc_clipsim}, we can notice how 
CLIP-similarity seems unable to recognize certain geometric and color details, present in the shapes generated by Liu et al.~\cite{liu2022towards}, which are critical for determining the shape most coherent to the given text prompt. 
In the first row of Figure~\ref{fig:quali_cc_clipsim} CLIP-Similarity prefers the table in the middle although it lacks the \emph{``grey cubicle base''} and the \emph{``tear drop shape''} described in the text. 
In the second row, CLIP-Similarity is not able to localize the parts containing the colors specified in the text. As discussed in Section~\ref{subsec:expres_HST}, such weakness may be caused by the difficulty of CLIP in processing fine-grained text descriptions and understanding the spatial relations among the objects' parts.

\begin{figure}[t!]
    \centering
    \includegraphics[width=0.97\linewidth]{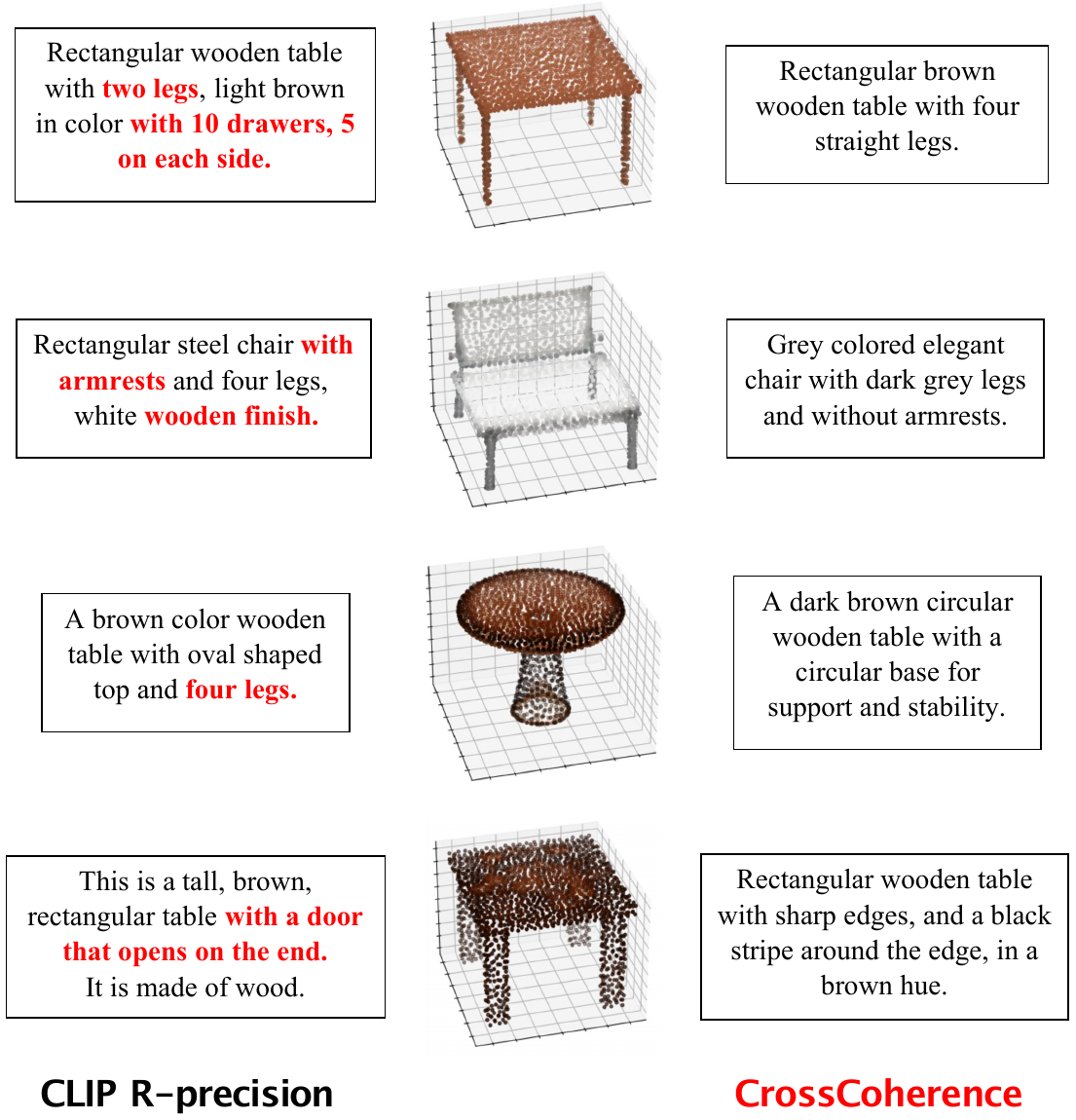}
    \caption{Examples of shapes generated by Shap-E with the corresponding text retrieved by CLIP R-precision, on the right, and \themetric{}, on the left.}
    \label{fig:quali_cc_rprecision}
\end{figure}

In Figure~\ref{fig:quali_cc_rprecision} we report some qualitative examples to motivate the large gap between \themetric{} and CLIP R-Precision. Here, it is possible to observe that even if in some cases CLIP R-precision retrieves the ground-truth text as the most coherent with the shape from Shap-E, these descriptions contain details that are not present in the generated shape (e.g., \emph{``two legs''} in the first row, \emph{``with armrests''} in the second row, \emph{``four legs''} in the third row and \emph{``with a door that opens on the end''} in the last row).
This behavior arises because Shap-E fails to generate all the specified details in the ground-truth text and CLIP R-precision cannot accurately recognize the absence of these elements in the generated shape.
On the contrary, \themetric{} is able to retrieve more comprehensive and accurate text prompts which are more coherent with the generated shapes.
In the supplementary, we provide additional qualitative examples. 


\section{Conclusions, Limitations and Future Work}
In this work, we presented the first benchmark for text-to-shape coherence, which includes a paired dataset of shapes and texts, \textbf{GPT2Shape}, a shape-text coherence quantitative metric, \textbf{\themetric{}}, and a human-validated test set, \textbf{HST}. Through extensive comparisons with existing works and a  user study, we have quantitatively demonstrated the superior quality of our dataset and evaluation metric.
Our benchmark may enable the field of text-driven shape generation to perform accurate comparisons along the important dimension of text-to-shape coherence.

Our results suggest that caution should be taken when using metrics that rely on rendered views to rank generative models, such as CLIP-based metrics, as they exhibit failures due to their inability to capture intricate nuances present in both textual descriptions and 3D shapes as well as a strong dependency on the rendering parameters. 

While we have demonstrated the potential of leveraging a LLM to improve the quality of the descriptions, it is important to note that our approach did not incorporate the shape as an additional input to the LLM itself. To further enhance descriptions and expand the benchmark beyond the ShapeNet categories covered in Text2Shape, one promising avenue is to explore the use of a Visual-Question-Answering models on 3D shape renderings. This extension may enable even more meaningful descriptions and address a limitation in our current work.


Finally, in light of this limitation, the proposed strategy of text refinement as well as the evaluation metric may be readily extended to a wider set of 3D objects.
Interestingly, two contemporary publications~\cite{cap3d, ulip2}, released during the writing of this manuscript, proposed methods to build large-scale paired text-shape datasets, which we plan to leverage to further develop our work. \\
\textbf{Acknowledgements}. We acknowledge the CINECA award under the ISCRA initiative, for the availability of high performance computing resources and support.

{\small
\bibliographystyle{ieee_fullname}
\bibliography{egbib}
}

\newpage\phantom{Supplementary}
\multido{\i=1+1}{16}{
\includepdf[pages={\i}]{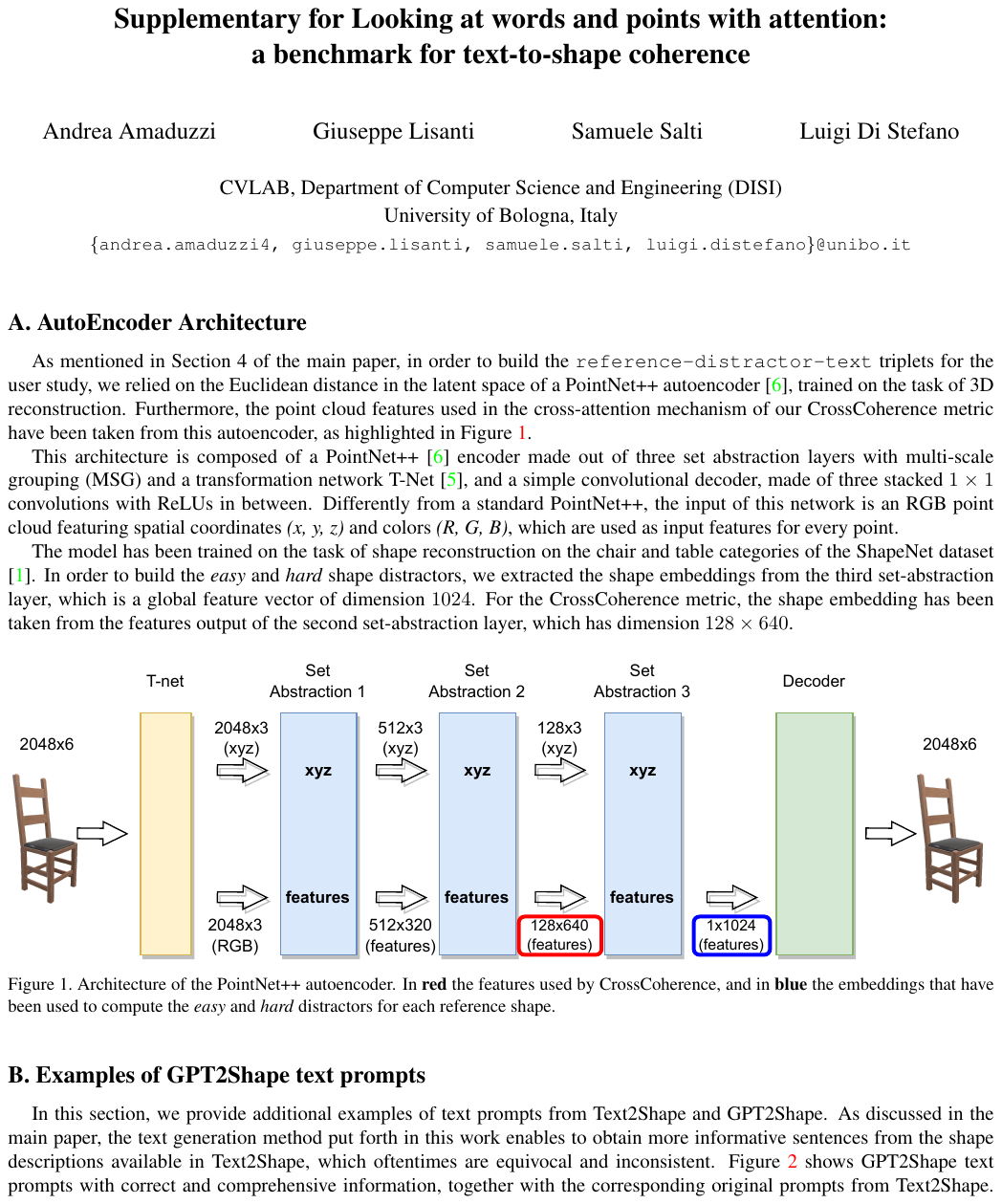}
}

\end{document}